\documentclass{article}
\usepackage{spconf,amsmath,graphicx,amssymb}
\usepackage{multicol}
\usepackage{siunitx}

\usepackage{setspace}

\usepackage{pifont}

\usepackage{paralist}
\usepackage{mathtools}

\usepackage[utf8]{inputenc}
\usepackage{pgfplots}
\usepackage{subcaption} 
\usepackage{tikz}
\usepackage{etoolbox}
\usepackage[outline]{contour}
\makeatletter
\patchcmd{\Ginclude@eps}{"#1"}{#1}{}{}
\makeatother
\usetikzlibrary{fit,calc,trees,positioning,arrows,chains,shapes.geometric,decorations.pathreplacing,decorations.pathmorphing,shapes,matrix,shapes.symbols,matrix,backgrounds,spy,calc}
\DeclareSIUnit{\mio}{Mio.}

\usepackage{makecell}

\title{Graph-based Motion Planning for automated Vehicles using multi-model branching and admissible heuristics}

\name{Oliver Speidel, Jona Ruof, and Klaus Dietmayer}
\address{\textit{Institute of Measurement, Control and Microtechnology}\\
	\textit{Ulm University}, 89081 Ulm, Germany\\
	firstname.lastname@uni-ulm.de}
\begin{document}
\maketitle

\begin{abstract}
Automated driving in urban scenarios requires efficient planning algorithms
able to handle complex situations in real-time. 
A popular approach is to use graph-based planning methods in order to obtain a rough trajectory which is subsequently optimized. A key aspect is the generation of trajectories implementing comfortable and safe behavior already during graph-search, while keeping computation times low.
To capture this aspect, on the one hand, a branching strategy is presented in this work that leads to better performance in terms of quality of resulting trajectories and runtime.
On the other hand, admissible heuristics are shown which guide the graph-search efficiently, where
the solution remains optimal.
\end{abstract}
\section{Introduction}
\label{sec:intro}
Over the last years, intensive research has been carried out in the field of autonomous driving \cite{Urmson2008,Ziegler2014a,Kunz2015}. 
Thereby, motion planning is a crucial requirement and one of the most challenging aspects for automated vehicles.
As early as 2007, impressive automated systems for complex urban scenarios with interacting vehicles were presented as part of the well-known Urban Challenge initiated by the Defense Advanced Research Projects Agency (DARPA) \cite{Urmson2008}.
In 2013, the Mercedes S-Class Bertha was able to drive fully autonomously more than 100km from Mannheim to Pforzheim in Germany \cite{Ziegler2014a, Ziegler2014}. 

A popular architecture for the motion planning system follows the idea that a behavior planning module decides for a strategic maneuver option, which is passed to a trajectory planning module where a feasible trajectory is calculated. 
A practicable approach for behavior planning is to generate a maneuver option rule-based using heuristics. However, this limits the capabilities for foresighted motion planning in complex environments \cite{Ziegler2015}.

For this reason more foresighted but still efficient behavior and trajectory planning systems are widely investigated. 
A popular concept for graph-based behavior planning is shown in \cite{Hubmann2016}. A speed profile along a given reference path is obtained by using graph-search methods. Therefore, a graph is generated where nodes represent states and edges represent actions. The idea is to extract a rough behavior trajectory over a planning horizon $t_\text{hor} \approx \SI{10}{\second}$ in order to enable foresighted behavior planning. The action set consists of discrete acceleration values and the temporal discretization is $\Delta t = \SI{1}{\second}$.
There exist several approaches which extend this concept of behavior planning for, e.g., short term lateral motion \cite{Zhan2017},  merging behavior at highways \cite{Ward2018a} or courteous behavior at intersections \cite{Speidel2019}.

In \cite{Zhang2020}, closed-loop forward simulation implementing high-level policies is used in order to generate subsequent states in the graph, in contrast to discrete acceleration or velocity values.
The forward simulation is done using the Intelligent Driver Model (IDM) \cite{Treiber2000} and the Pure Pursuit Controller \cite{Coulter1992}. However, due to the computational complexity of the approach, the concept is restricted to a horizon of $t_\text{hor} = \SI{8}{\second}$ seconds and large discretization of $\Delta t = \SI{2}{\second}$ . Further, only one policy change is allowed within the planning horizon $t_\text{hor}$. 
A similar method is used in \cite{Lenz2016} where cooperative behavior for highway scenarios is generated using Monte Carlo Tree Search. In this concept, the IDM as well as pre-defined acceleration actions are employed.
Driver models could also be successfully used in various other concepts to efficiently generate social compliant behavior.
For example, in \cite{Graf2019} the IDM-based MOBIL model \cite{Kesting2007} is utilized in order to decide whether a lane change is desirable. 

Based on the previous discussion, in this work, a motion planning framework is developed enabling foresighted and courteous behavior using graph-search methods extending our concept presented in \cite{Speidel2019}.
The main idea is to utilize different control and driver models, which are known to generate preferable actions for specific scenarios.
Consequently, we are able to improve the performance of graph-based behavior planning and driven trajectories compared to related work \cite{Hubmann2016, Speidel2019}.
In order to still assure real-time capabilities and significantly reduce calculation times, we propose action selection strategies as well as efficient admissible heuristics, which are applicable in interactive urban scenarios.

\section{Methodology}
\label{sec:methodology}
\begin{figure}
	\centering
		\begin{tikzpicture}
	\definecolor{compcolor}{RGB}{224,224,224};
	
	\tikzstyle{section} = [fill=compcolor, draw,rectangle,rounded corners,minimum width=2.25cm,minimum height=1.5cm, align=left,text depth = 1.5cm, draw,anchor=north, inner sep=0.1cm];
	
	\tikzstyle{subsection} = [fill=white,draw,anchor=south,rectangle,rounded corners,dashed,minimum width=2.0cm,minimum height=0.5cm, align=left];
	

	\node[section, minimum width=3.9cm, minimum height=2.25cm, text depth = 1.5cm, label={[shift={(18.5ex,-6.25ex)}, align=left]north west: Graph-based \\ Behavior Planning}] (BehaviorPlanning) at (-1.75,0.0) {};
	\node[subsection,shift={(0,0.8)}, minimum width=3.7cm] (innerboxone) at (BehaviorPlanning.south){\footnotesize {Branching Strategy (Sec. \ref{subsec:contextAwareBranching})}};
	\node[subsection,shift={(0,-0.65)}] (innerboxtwo) at (innerboxone.south){\footnotesize {Costs and Heuristics (Sec. \ref{subsec:heuristicFunctions})}};
	\node[section, minimum width=1.5cm,text depth = 0.0cm, minimum height=1.0cm] (TrajectoryPlanning) at (1.5,-0.65) {Trajectory\\Planning};

	\node[fit=(BehaviorPlanning)(TrajectoryPlanning),dashed,draw,rounded corners,inner sep=0.4cm, shift={(0,0.2)}, label={[shift={(25ex,-4ex)}]north west: Motion Planning (Sec. \ref{sec:methodology})}](MotionPlanning){};
	
	\node[rectangle,rounded corners,fill=compcolor,draw,xshift=-0.675cm] at(MotionPlanning.west)(envdata){\rotatebox[]{90} {Environmental Data}};

	\node[rectangle,rounded corners,fill=compcolor,draw,xshift=0.675cm]
	at(MotionPlanning.east)(controller){\rotatebox[]{90} {Controller}};	
	
	\path[-latex',draw, line width=0.75pt](MotionPlanning)--(controller);
	\path[-latex',draw, line width=0.75pt](envdata)--(MotionPlanning);
	\path[-latex',draw, line width=0.75pt](BehaviorPlanning)--(TrajectoryPlanning);
	
	\end{tikzpicture}
	\vspace*{-0.5cm}
	\caption{System Overview}
	\label{fig:Architecture}
\end{figure}
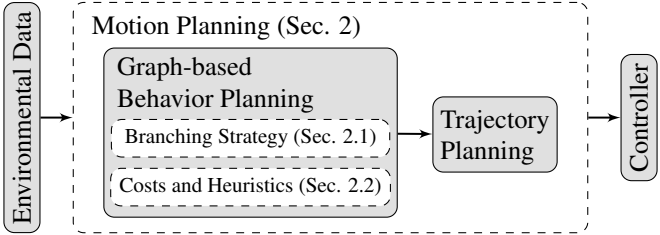

The concept follows the modular architecture of behavior and trajectory planning as shown in Figure~\ref{fig:Architecture}. Preceding modules provide environmental data including map data as well as state estimations of other vehicles with according predictions. Thereby, a set of predicted trajectories for each other vehicle with corresponding uncertainties is received.

The goal of the graph-based behavior planning is to obtain a rough behavior trajectory. In general, planning is done relative to the center line of the current road lane. Therefore, a node in the graph is represented by a state vector
\begin{equation}
	\boldsymbol{x}_k = [s_k, d_k, \theta_k, \kappa_k, v_k, a_k]^\mathsf{T}, \quad k \in [0,\dots, T] \,,
\end{equation}
where $s$ is the longitudinal position along the lane, $d$ the lateral distance to the lane, $\theta$ the orientation, $\kappa$ the curvature, $v$ the velocity, $a$ the acceleration and $k$ the corresponding time step. The end of the planning horizon $t_\text{hor}$ is denoted by the index $T$. The lane relative position $[s,d]$ can be transformed to the classic representation $[x,y]$ in Cartesian coordinates and vice versa. For further details, the reader is referred to \cite{Werling2010}.

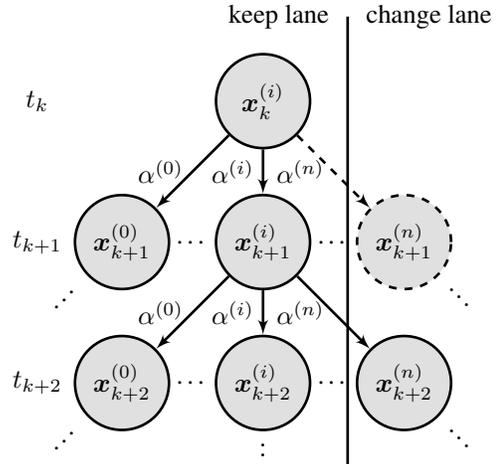
\begin{figure}
	\centering
		\begin{tikzpicture}
	\definecolor{compcolor}{RGB}{224,224,224};
	\def\xsh{0.75cm};
	\def\ysh{0.75cm};
	\def\ashift{0.15cm};
	\def\tstepone{k};
	\def\tsteptwo{k+1};
	\def\tstepthree{k+2};
	\def\actionone{(0)};
	\def\actiontwo{(i)};
	\def\actionthree{(n)};
	\def\lw{1pt};
	\tikzstyle{treenode} = [circle,draw,minimum size=1.25cm,line width=\lw,fill=compcolor];
	\tikzstyle{treepath} = [-latex',draw,line width=\lw];

	\node[treenode] at (0,0)(x1){$\boldsymbol x^{\actiontwo}_{\tstepone}$};
	\node[] at (-4*\xsh, 0*\ysh)(t0){$t_{\tstepone}$};
	
	\node[treenode] at (-2.5*\xsh, -2.5*\ysh)(x2){$\boldsymbol x^{\actionone}_{\tsteptwo}$};
	\node[treenode] at (0*\xsh, -2.5*\ysh)(x3){$\boldsymbol x^{\actiontwo}_{\tsteptwo}$};
	\node[treenode, dashed] at (2.5*\xsh, -2.5*\ysh)(x4){$\boldsymbol x^{\actionthree}_{\tsteptwo}$};
	\node[ rotate=45] at (-3.5*\xsh, -3.5*\ysh)(pointsx2){\dots};
	\node[ rotate=-45] at (3.5*\xsh, -3.5*\ysh)(pointsx4){\dots};
	\node (pointsx2x3) at ($(x2)!0.5!(x3)$) {\dots};
	\node (pointsx3x4) at ($(x3)!0.5!(x4)$) {\dots};
	\path[treepath](x1)--(x2) node [midway, shift={(\ashift,0)}, label=left:{$\alpha^{\actionone}$}] {};
	\path[treepath](x1)--(x3) node [midway, shift={(\ashift,0)}, label=left:{$\alpha^{\actiontwo}$}] {};
	\path[treepath, dashed](x1)--(x4)  node [midway, shift={(\ashift,0)}, label=left:{$\alpha^{\actionthree}$}] {};
	\node[] at (-4*\xsh, -2.5*\ysh)(t1){$t_{\tsteptwo}$};

	\node[treenode] at (-2.5*\xsh, -5*\ysh)(x5){$\boldsymbol x^{\actionone}_{\tstepthree}$};
	\node[treenode] at (0*\xsh, -5*\ysh)(x6){$\boldsymbol x^{\actiontwo}_{\tstepthree}$};
	\node[treenode] at (2.5*\xsh, -5*\ysh)(x7){$\boldsymbol x^{{\actionthree}}_{\tstepthree}$};
	
	\node[ rotate=45] at (-3.5*\xsh, -6*\ysh)(pointsx2){\dots};
	\node[ rotate=-45] at (3.5*\xsh, -6*\ysh)(pointsx4){\dots};
	\node[ rotate=90] at (0.0*\xsh, -6.25*\ysh)(pointsx4){\dots};
	\node (pointsx2x3) at ($(x5)!0.5!(x6)$) {\dots};
	\node (pointsx3x4) at ($(x6)!0.5!(x7)$) {\dots};
	
	\path[treepath](x3)--(x5) node [midway, shift={(\ashift,0)}, label=left:{$\alpha^{\actionone}$}] {};
	\path[treepath](x3)--(x6) node [midway, shift={(\ashift,0)}, label=left:{$\alpha^{\actiontwo}$}] {};
	\path[treepath](x3)--(x7)  node [midway, shift={(\ashift,0)}, label=left:{$\alpha^{\actionthree}$}] {};
	
	\node[] at (-4*\xsh, -5.0*\ysh)(t1){$t_{\tstepthree}$};
	
	\coordinate (lineBegin) at (1.5*\xsh,-6.5*\ysh);
	\coordinate (lineEnd) at (1.5*\xsh,1.5*\ysh);
	\draw[draw,line width=\lw](lineBegin)--(lineEnd) node {} ;
	\node[label=right:change lane] at (lineEnd) {};
	\node[label=left:keep lane] at (lineEnd) {};
	\end{tikzpicture}
	\vspace*{-0.4cm}
	\caption{Exemplary part of the utilized graph structure for behavior planning. Nodes represent states connected by edges associated with actions contained in \mbox{$\mathcal{A} = \{\alpha^{(0)},\dots, \alpha^{(n)}\}$}. The concept of MOBIL-based action selection is demonstrated for a possible lane change maneuver, which is not expanded originating from $\boldsymbol x_k^{(i)}$ indicated by dashed lines. However, a possible lane change is regarded originating from $\boldsymbol x^{(i)}_{k+1}$.}
	\label{fig:graph}
\end{figure}

The expansion of a node, i.e. the generation of possible subsequent states, is done using different models which will be presented in Section~\ref{subsec:contextAwareBranching}. The time discretization of subsequent nodes is $\Delta t = \SI{1}{\second}$. Beginning from the root node, i.e. the current state, a graph is generated up to the planning horizon $t_\text{hor} \approx 10s$. The graph structure is exemplary shown in Figure~\ref{fig:graph}.
The optimal behavior trajectory is extracted using the A*-search algorithm, where the search is guided by the admissible heuristic functions shown in Section~\ref{subsec:heuristicFunctions}.
The generation of the resulting trajectory in the trajectory planning module is based on the approach presented in \cite{Speidel2019}.
The general idea is to use polynomials in order to interpolate between the behavior trajectory states.
In contrast to our previous work, in this work we also regard lateral optimization. 
In the end, the resulting trajectory is passed to the controller which generates the input for the actuators. 

\subsection{Branching Strategy}
\label{subsec:contextAwareBranching}

The idea of the branching strategy is to combine the advantages of pre-defined acceleration actions and model-based action which generate preferable behavior for different scenarios, inspired by \cite{Lenz2016}.
In order to omit the expansion of all actions, only a subset of actions is expanded at each node. In the following, this process of choosing which action to expand at which node is also referred to as action selection. 
In general, the ideas of \cite{Lenz2016} are extended by additional control models as well as more sophisticated action selection strategies.
Further, in this work, the behavior planning is embedded into a holistic framework generating comfortable trajectories. In addition, the solution of behavior planning guarantees the existence of a feasible solution in the trajectory planning module, as all kinematic and collision constraints are considered during forward simulation.

In general, longitudinal actions $\mathcal{A}_\text{lon}$ and lateral actions $\mathcal{A}_\text{lat}$ can be distinguished, where the resulting action set is $\mathcal{A} = \mathcal{A}_\text{lon} \times \mathcal{A}_\text{lat} = \{\alpha^{(0)},\dots, \alpha^{(n)}\}$.
Hereafter, the different actions and corresponding action selection strategies are presented. 
\newline
\textbf{Longitudinal actions:} 
For longitudinal action generation, acceleration and velocity targets are distinguished .

First the acceleration targets are discussed. These consist of pre-defined accelerations $a^{(i)} \in \{-2, -1, 0, 1, 2\} \SI{}{\meter\per\second\squared}$ and the acceleration according to the IDM $a^{\text{(idm)}}$. The $a^{\text{(idm)}}$ is expanded during car-following scenarios, as it is able to model comfortable and human-like following behavior. In order to omit expansions of similar states, in car-following scenarios only generic acceleration targets $a^{(i)}$ with \mbox{$|a^{(i)} - a^{\text{(idm)}}| > \SI{0.5}{\meter\per\second\squared}$} are expanded. Further, we require \mbox{$|a_{k+1} - a_k| \leq 1.9 \SI{}{\meter\per\second\squared}$}.

The longitudinal state transition model for acceleration targets is defined by
\begin{multline}
\setlength\arraycolsep{1.5pt}
\begin{bmatrix}
	s_{k+1} \\
	v_{k+1} \\
	a_{k+1}
	\end{bmatrix}
=
\begin{bmatrix}
1 & \Delta t & \frac{1}{2}\Delta t^2\\ 
0 & 1 & \Delta t\\ 
0 & 0 & 1\\
\end{bmatrix}
\begin{bmatrix}
s_k \\
v_k \\
a_k \\
\end{bmatrix}
+
\begin{bmatrix}
\frac{1}{6}\Delta t^3\\[1pt]
\frac{1}{2}\Delta t^2 \\
\Delta t
\end{bmatrix}
\dot{a}_k\,,
\label{eqn:caTransition}
\end{multline}
where $\dot{a}_k = (a^{(i)} - a_{k})/ \Delta t$ and, as a result, $a_{k+1} = a^{(i)}$. 

The velocity targets are defined by the desired velocity $v_d$ and stillstand $v_0 = 0$, where the acceleration of the targets is constrained to 0. The expansion of these actions is triggered if the target velocity is reachable within $\Delta t$. As state transition model, the concept of C1-continuous time optimal trajectories summarized in \cite{Knierim2012} is employed. Thus, comfort is ensured by restricted and continuous jerk. In general, C1-continuous time optimal trajectories also allow emergency maneuvers at kinematic limits.  However, in this work, we  limit our scope to non-safety critical scenarios.
For further details, the reader is referred to \cite{Knierim2012}.
\newline
\textbf{Lateral actions:} The lateral action set is given by the different road lanes which can be targeted to drive on. Therefore, $\mathcal{A}_\text{lat} = \{r_\text{l}, r_\text{c}, r_\text{r}\}$ where $r_\text{l}$ represents the lane to the left, $r_\text{c}$ the current lane and $r_\text{r}$ the lane to the right. Using $r_c$ the motion can be modeled purely longitudinal along the center line of the current lane.
In order to perform a lane change to $r_\text{l}$ or $r_\text{r}$, the Pure Pursuit Controller is employed as lateral transition model 
and, consequently, the vehicle state is regarded in Cartesian Coordinates. 
The steering behavior defined by the Pure Pursuit Controller is combined with different acceleration targets for longitudinal behavior.
This allows to restrict $\kappa$, $\dot{\kappa}$ and the absolute acceleration $a_{\text{abs}}$ already during behavior planning which ensures feasible solutions in the trajectory planning module. 
The context in which a lane change is explored is defined by the MOBIL model \cite{Kesting2007}, which is known to generate human-like decision-making for lane change behavior \cite{Graf2019}. Thereby, it is estimated if a lane change is favorable for the combined costs of all involved vehicles.

\subsection{Cost and Heuristic Functions}
\label{subsec:heuristicFunctions}

The costs attributed to a node are defined by
\begin{equation}
\vspace{-0.1cm}
	J = w_\text{f} j_\text{f} + w_\text{c} j_\text{c} + w_v j_v + w_a j_a + w_{\dot{a}} j_{\dot{a}} + w_{\text{lc}} j_\text{lc}\,,
\end{equation}
where $j_\text{f}$ represents costs for the spatio-temporal distance to the vehicle in front, $j_\text{c}$ are courtesy costs that arise if the ego vehicle pulls out or drives in front of another vehicle \cite{Speidel2019}. The difference to the desired velocity is regarded by $j_v$ and the comfort is optimized by costs $j_a$ and $j_{\dot{a}}$ for larger absolute values of $a$ and $\dot{a}$. Further, costs $j_\text{lc}$ arise for lane changes. The single cost terms can be weighted with the according cost weighting $w$. In order to generate courteous and safe behavior, a set of predicted trajectories for each other vehicle is considered, where the corresponding uncertainties are incorporated by the single cost terms.

To further improve the runtime, admissible heuristics are developed which can be calculated online.
The idea is to use a linear combination of heuristic terms rather than model directly one overall heuristic. 
If the heuristics for the single cost terms are admissible, the linear combination remains admissible \cite{Russell2016}.
Therefore, the heuristic is given with 
\begin{equation}
h_{\text{all}} = \sum_{i=k+1}^{T} w_\text{f} j_{\text{f},i}^\text{min} + w_\text{c} j_{\text{c},i}^\text{min} + w_v j_{v,i}^\text{min} + w_a j_{a,i}^\text{min} + w_{\dot{a}} j_{\dot{a},i}^\text{min} \,,
\end{equation}
where $j^\text{min}_{(\cdot),i}$ represent the minimal costs for the corresponding cost term that arise at time $i$ originating from the currently expanded node $\boldsymbol{x}_k$. In the following, the calculation of the single minimal costs terms is explained. The terms $ j_{\dot{a},i}^\text{min}$ and $ j_{a,i}^\text{min}$ are determined by the minimal necessary jerk and acceleration to avoid a collision with the vehicle in front. The term $j_{\text{f},i}^\text{min}$ can be estimated by calculating the maximum possible distance to the vehicle in front at $i$. The same applies for $j_{\text{c},i}^\text{min}$, where the maximum possible distance to the vehicle behind is calculated. The minimal arising velocity costs $j_{v,i}^\text{min}$ are given if the ego vehicle accelerates with maximum acceleration to the desired speed. Even though the single minimal costs terms result in low estimated heuristic costs, the evaluation shows that the combination of all heuristics leads to a significant reduction of calculation times, while the solution remains optimal.
\begin{figure*}[ht]
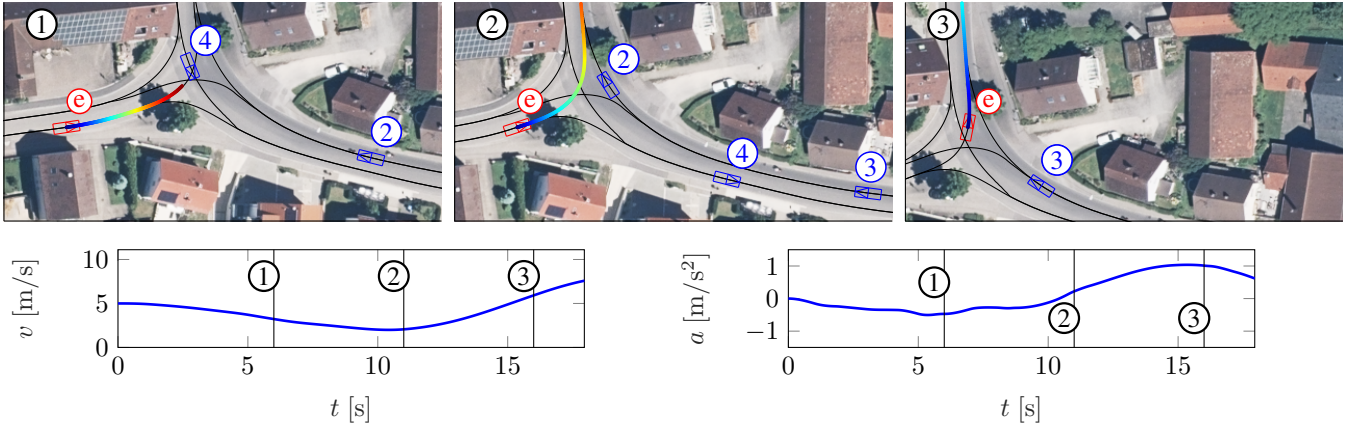

	\setlength\columnsep{5pt}
	\begin{multicols}{3}
		\vspace{-0.65cm}
		\input{eval_result1.tikz}\;
		\input{eval_result2.tikz}\;
		\input{eval_result3.tikz}
	\end{multicols}
\vspace{-0.675cm}
	\begin{multicols}{2}
%
%
\begin{tikzpicture}

\begin{axis}[%
width=6.2cm,
height=1.3cm,
at={(0cm,0cm)},
scale only axis,
xmin=0,
xmax=17.9500000000001,
xlabel style={font=\color{white!15!black}},
xlabel={$t\;[\SI[per-mode=repeated-symbol]{}{\second}]$},
ymin=0,
ymax=11.1111111111111,
ylabel style={font=\color{white!15!black}},
ylabel={$v\;[\SI[per-mode=repeated-symbol]{}{\metre\per\second}]$},
axis background/.style={fill=white}
]
\addplot [color=blue, line width=1.0pt, forget plot]
  table[row sep=crcr]{%
0	5\\
0.05	4.99999\\
0.1	4.99993\\
0.15	4.99976\\
0.2	4.99945\\
0.25	4.99893\\
0.3	4.99818\\
0.35	4.99715\\
0.4	4.99581\\
0.45	4.99412\\
0.5	4.99204\\
0.55	4.98956\\
0.6	4.98664\\
0.65	4.98326\\
0.7	4.97939\\
0.75	4.97502\\
0.8	4.97012\\
0.85	4.96469\\
0.9	4.95869\\
0.95	4.95214\\
1	4.945\\
1.05	4.93728\\
1.1	4.92903\\
1.15	4.92028\\
1.2	4.91109\\
1.25	4.90149\\
1.3	4.89152\\
1.35	4.88123\\
1.4	4.87065\\
1.45	4.8598\\
1.5	4.84874\\
1.55	4.83747\\
1.6	4.82604\\
1.65	4.81447\\
1.7	4.80278\\
1.75	4.791\\
1.8	4.77915\\
1.85	4.76723\\
1.9	4.75522\\
1.95	4.74309\\
2	4.73081\\
2.05	4.71834\\
2.1	4.70568\\
2.15	4.69283\\
2.2	4.67979\\
2.25	4.66656\\
2.3	4.65313\\
2.35	4.63951\\
2.4	4.6257\\
2.45	4.61169\\
2.5	4.59748\\
2.55	4.58308\\
2.6	4.56848\\
2.65	4.55368\\
2.7	4.53869\\
2.75	4.5235\\
2.8	4.50812\\
2.85	4.49254\\
2.9	4.47677\\
2.95	4.46081\\
3	4.44466\\
3.05	4.42832\\
3.1	4.41182\\
3.15	4.39517\\
3.2	4.37839\\
3.25	4.3615\\
3.3	4.34452\\
3.35	4.32747\\
3.4	4.31034\\
3.45	4.29317\\
3.5	4.27596\\
3.55	4.25873\\
3.6	4.24148\\
3.65	4.22422\\
3.69999999999999	4.20694\\
3.74999999999999	4.18966\\
3.79999999999999	4.17235\\
3.84999999999999	4.15503\\
3.89999999999999	4.13768\\
3.94999999999999	4.12031\\
3.99999999999999	4.10291\\
4.04999999999999	4.08547\\
4.09999999999999	4.06801\\
4.14999999999999	4.0505\\
4.19999999999999	4.03294\\
4.24999999999999	4.01533\\
4.29999999999999	3.99761\\
4.34999999999999	3.97973\\
4.39999999999999	3.96163\\
4.44999999999999	3.94328\\
4.49999999999999	3.92462\\
4.54999999999999	3.90563\\
4.59999999999999	3.88627\\
4.64999999999999	3.8665\\
4.69999999999999	3.84631\\
4.74999999999999	3.82566\\
4.79999999999999	3.80454\\
4.84999999999999	3.78292\\
4.89999999999999	3.7608\\
4.94999999999999	3.73816\\
4.99999999999999	3.71499\\
5.04999999999999	3.69131\\
5.09999999999999	3.66718\\
5.14999999999999	3.64269\\
5.19999999999999	3.61791\\
5.24999999999999	3.59293\\
5.29999999999999	3.5678\\
5.34999999999999	3.54259\\
5.39999999999999	3.51737\\
5.44999999999999	3.49219\\
5.49999999999999	3.4671\\
5.54999999999999	3.44215\\
5.59999999999999	3.4174\\
5.64999999999999	3.39287\\
5.69999999999999	3.36855\\
5.74999999999999	3.34443\\
5.79999999999999	3.3205\\
5.84999999999999	3.29673\\
5.89999999999999	3.27308\\
5.94999999999999	3.24949\\
5.99999999999999	3.22594\\
6.04999999999999	3.2024\\
6.09999999999999	3.17891\\
6.14999999999999	3.15552\\
6.19999999999999	3.13229\\
6.24999999999999	3.10927\\
6.29999999999999	3.08649\\
6.34999999999999	3.064\\
6.39999999999999	3.04184\\
6.44999999999999	3.02002\\
6.49999999999998	2.9986\\
6.54999999999998	2.97758\\
6.59999999999998	2.95701\\
6.64999999999998	2.9369\\
6.69999999999998	2.91727\\
6.74999999999998	2.89815\\
6.79999999999998	2.87954\\
6.84999999999998	2.86147\\
6.89999999999998	2.84393\\
6.94999999999998	2.82694\\
6.99999999999998	2.81049\\
7.04999999999998	2.79458\\
7.09999999999998	2.77913\\
7.14999999999998	2.76407\\
7.19999999999998	2.74932\\
7.24999999999998	2.73482\\
7.29999999999998	2.72052\\
7.34999999999998	2.7064\\
7.39999999999998	2.69243\\
7.44999999999998	2.67856\\
7.49999999999998	2.66479\\
7.54999999999998	2.65107\\
7.59999999999998	2.63741\\
7.64999999999998	2.62376\\
7.69999999999998	2.61013\\
7.74999999999998	2.59649\\
7.79999999999998	2.58283\\
7.84999999999998	2.56914\\
7.89999999999998	2.55541\\
7.94999999999998	2.54164\\
7.99999999999998	2.52782\\
8.04999999999998	2.51394\\
8.09999999999998	2.49998\\
8.14999999999998	2.4859\\
8.19999999999998	2.47166\\
8.24999999999998	2.45726\\
8.29999999999998	2.44271\\
8.34999999999998	2.42804\\
8.39999999999998	2.41328\\
8.44999999999999	2.39847\\
8.49999999999999	2.38365\\
8.54999999999999	2.36889\\
8.59999999999999	2.35424\\
8.64999999999999	2.33972\\
8.69999999999999	2.32533\\
8.74999999999999	2.31102\\
8.79999999999999	2.29676\\
8.84999999999999	2.28253\\
8.89999999999999	2.26836\\
8.94999999999999	2.25427\\
8.99999999999999	2.24029\\
9.04999999999999	2.22645\\
9.09999999999999	2.21278\\
9.15	2.1993\\
9.2	2.18604\\
9.25	2.17302\\
9.3	2.16026\\
9.35	2.1478\\
9.4	2.13565\\
9.45	2.12384\\
9.5	2.11239\\
9.55	2.10133\\
9.6	2.09069\\
9.65	2.08048\\
9.7	2.07073\\
9.75	2.06146\\
9.8	2.0527\\
9.85	2.04447\\
9.90000000000001	2.03679\\
9.95000000000001	2.02968\\
10	2.02316\\
10.05	2.01727\\
10.1	2.012\\
10.15	2.0074\\
10.2	2.00348\\
10.25	2.00025\\
10.3	1.99777\\
10.35	1.99609\\
10.4	1.99523\\
10.45	1.99525\\
10.5	1.99616\\
10.55	1.99801\\
10.6	2.00081\\
10.65	2.00459\\
10.7	2.00937\\
10.75	2.01517\\
10.8	2.02201\\
10.85	2.02988\\
10.9	2.03875\\
10.95	2.04855\\
11	2.05921\\
11.05	2.0707\\
11.1	2.08298\\
11.15	2.09603\\
11.2	2.10984\\
11.25	2.12439\\
11.3	2.13967\\
11.35	2.15566\\
11.4	2.17234\\
11.45	2.18971\\
11.5	2.20773\\
11.55	2.22637\\
11.6	2.24562\\
11.65	2.26545\\
11.7	2.28586\\
11.75	2.30687\\
11.8	2.32848\\
11.85	2.35071\\
11.9	2.37356\\
11.95	2.39703\\
12	2.42114\\
12.05	2.44588\\
12.1	2.47127\\
12.15	2.49731\\
12.2	2.52399\\
12.25	2.55133\\
12.3	2.57931\\
12.35	2.60796\\
12.4	2.63725\\
12.45	2.66721\\
12.5	2.69781\\
12.55	2.72906\\
12.6	2.76096\\
12.65	2.7935\\
12.7	2.82668\\
12.75	2.86051\\
12.8	2.89496\\
12.85	2.93004\\
12.9	2.96574\\
12.95	3.00207\\
13	3.039\\
13.0500000000001	3.07655\\
13.1000000000001	3.11469\\
13.1500000000001	3.15343\\
13.2000000000001	3.19275\\
13.2500000000001	3.23265\\
13.3000000000001	3.2731\\
13.3500000000001	3.3141\\
13.4000000000001	3.35564\\
13.4500000000001	3.39769\\
13.5000000000001	3.44025\\
13.5500000000001	3.4833\\
13.6000000000001	3.52685\\
13.6500000000001	3.57087\\
13.7000000000001	3.61536\\
13.7500000000001	3.6603\\
13.8000000000001	3.70568\\
13.8500000000001	3.75149\\
13.9000000000001	3.79771\\
13.9500000000001	3.84433\\
14.0000000000001	3.89132\\
14.0500000000001	3.93868\\
14.1000000000001	3.98638\\
14.1500000000001	4.03443\\
14.2000000000001	4.08279\\
14.2500000000001	4.13147\\
14.3000000000001	4.18043\\
14.3500000000001	4.22968\\
14.4000000000001	4.27919\\
14.4500000000001	4.32894\\
14.5000000000001	4.37893\\
14.5500000000001	4.42914\\
14.6000000000001	4.47954\\
14.6500000000001	4.53013\\
14.7000000000001	4.58088\\
14.7500000000001	4.63179\\
14.8000000000001	4.68283\\
14.8500000000001	4.73398\\
14.9000000000001	4.78525\\
14.9500000000001	4.83662\\
15.0000000000001	4.88808\\
15.0500000000001	4.93962\\
15.1000000000001	4.99124\\
15.1500000000001	5.04292\\
15.2000000000001	5.09465\\
15.2500000000001	5.14642\\
15.3000000000001	5.19822\\
15.3500000000001	5.25003\\
15.4000000000001	5.30185\\
15.4500000000001	5.35365\\
15.5000000000001	5.40543\\
15.5500000000001	5.45717\\
15.6000000000001	5.50885\\
15.6500000000001	5.56046\\
15.7000000000001	5.61199\\
15.7500000000001	5.66343\\
15.8000000000001	5.71475\\
15.8500000000001	5.76595\\
15.9000000000001	5.81699\\
15.9500000000001	5.86786\\
16.0000000000001	5.91853\\
16.0500000000001	5.96899\\
16.1000000000001	6.01927\\
16.1500000000001	6.06939\\
16.2000000000001	6.11938\\
16.2500000000001	6.16925\\
16.3000000000001	6.21895\\
16.3500000000001	6.26844\\
16.4000000000001	6.31766\\
16.4500000000001	6.36657\\
16.5000000000001	6.4151\\
16.5500000000001	6.46323\\
16.6000000000001	6.5109\\
16.6500000000001	6.55809\\
16.7000000000001	6.60477\\
16.7500000000001	6.65093\\
16.8000000000001	6.69655\\
16.8500000000001	6.74163\\
16.9000000000001	6.78615\\
16.9500000000001	6.83009\\
17.0000000000001	6.87345\\
17.0500000000001	6.91623\\
17.1000000000001	6.95842\\
17.1500000000001	7.00005\\
17.2000000000001	7.04112\\
17.2500000000001	7.08165\\
17.3000000000001	7.12163\\
17.3500000000001	7.16102\\
17.4000000000001	7.19982\\
17.4500000000001	7.23801\\
17.5000000000001	7.27556\\
17.5500000000001	7.31247\\
17.6000000000001	7.34871\\
17.6500000000001	7.38428\\
17.7000000000001	7.41917\\
17.7500000000001	7.45335\\
17.8000000000001	7.48683\\
17.8500000000001	7.51959\\
17.9000000000001	7.5516\\
17.9500000000001	7.58285\\
};
\addplot [color=black, forget plot]
  table[row sep=crcr]{%
6	0\\
6	11.1111111111111\\
};
\node[right, align=left, circle,fill=white,draw=black,thick,line width=0.3mm,inner sep=1pt]
at (axis cs:5,8.111) {1};
\addplot [color=black, forget plot]
  table[row sep=crcr]{%
11	0\\
11	11.1111111111111\\
};
\node[right, align=left, circle,fill=white,draw=black,thick,line width=0.3mm,inner sep=1pt]
at (axis cs:10,8.111) {2};
\addplot [color=black, forget plot]
  table[row sep=crcr]{%
16	0\\
16	11.1111111111111\\
};
\node[right, align=left, circle,fill=white,draw=black,thick,line width=0.3mm,inner sep=1pt]
at (axis cs:15,8.111) {3};
\end{axis}
\end{tikzpicture}%
\;\;\;\;\;\;\;\;\;\;\;\;\columnbreak
%
%
\begin{tikzpicture}

\begin{axis}[%
width=6.2cm,
height=1.3cm,
at={(0cm,0cm)},
scale only axis,
xmin=0,
xmax=17.9500000000001,
xlabel style={font=\color{white!15!black}},
xlabel={$t\;[\SI[per-mode=repeated-symbol]{}{\second}]$},
ymin=-1.5,
ymax=1.5,
ylabel style={font=\color{white!15!black}},
ylabel={$a\;[\SI[per-mode=repeated-symbol]{}{\metre\per\second\squared}]$},
axis background/.style={fill=white}
]
\addplot [color=blue, line width=1.0pt, forget plot]
  table[row sep=crcr]{%
0	4.11978e-15\\
0.05	-0.000539307\\
0.1	-0.00211768\\
0.15	-0.00467689\\
0.2	-0.00816015\\
0.25	-0.0125121\\
0.3	-0.0176788\\
0.35	-0.0236077\\
0.4	-0.0302477\\
0.45	-0.037549\\
0.5	-0.0454632\\
0.55	-0.0539431\\
0.6	-0.0629431\\
0.65	-0.0724187\\
0.7	-0.0823267\\
0.75	-0.0926253\\
0.8	-0.103274\\
0.85	-0.114233\\
0.9	-0.125465\\
0.95	-0.136932\\
1	-0.1486\\
1.05	-0.159861\\
1.1	-0.170164\\
1.15	-0.179555\\
1.2	-0.188079\\
1.25	-0.195781\\
1.3	-0.202702\\
1.35	-0.208885\\
1.4	-0.214371\\
1.45	-0.219199\\
1.5	-0.223407\\
1.55	-0.227032\\
1.6	-0.230112\\
1.65	-0.232682\\
1.7	-0.234775\\
1.75	-0.236425\\
1.8	-0.237665\\
1.85	-0.23909\\
1.9	-0.241241\\
1.95	-0.24407\\
2	-0.247529\\
2.05	-0.2513\\
2.1	-0.255092\\
2.15	-0.258897\\
2.2	-0.262712\\
2.25	-0.266549\\
2.3	-0.27042\\
2.35	-0.27432\\
2.4	-0.278239\\
2.45	-0.282171\\
2.5	-0.286109\\
2.55	-0.290046\\
2.6	-0.293976\\
2.65	-0.297897\\
2.7	-0.301809\\
2.75	-0.30571\\
2.8	-0.309595\\
2.85	-0.313461\\
2.9	-0.317305\\
2.95	-0.321124\\
3	-0.324915\\
3.05	-0.32847\\
3.1	-0.331603\\
3.15	-0.334342\\
3.2	-0.33671\\
3.25	-0.338734\\
3.3	-0.340436\\
3.35	-0.34184\\
3.4	-0.342968\\
3.45	-0.343842\\
3.5	-0.344481\\
3.55	-0.344905\\
3.6	-0.345134\\
3.65	-0.345311\\
3.69999999999999	-0.345565\\
3.74999999999999	-0.345888\\
3.79999999999999	-0.346274\\
3.84999999999999	-0.346717\\
3.89999999999999	-0.347211\\
3.94999999999999	-0.34775\\
3.99999999999999	-0.348329\\
4.04999999999999	-0.348982\\
4.09999999999999	-0.349743\\
4.14999999999999	-0.350605\\
4.19999999999999	-0.351562\\
4.24999999999999	-0.353156\\
4.29999999999999	-0.355873\\
4.34999999999999	-0.359629\\
4.39999999999999	-0.364342\\
4.44999999999999	-0.369931\\
4.49999999999999	-0.376318\\
4.54999999999999	-0.383428\\
4.59999999999999	-0.391189\\
4.64999999999999	-0.399528\\
4.69999999999999	-0.408378\\
4.74999999999999	-0.417672\\
4.79999999999999	-0.427346\\
4.84999999999999	-0.437337\\
4.89999999999999	-0.447586\\
4.94999999999999	-0.458034\\
4.99999999999999	-0.468626\\
5.04999999999999	-0.478391\\
5.09999999999999	-0.486458\\
5.14999999999999	-0.492921\\
5.19999999999999	-0.49787\\
5.24999999999999	-0.501391\\
5.29999999999999	-0.50357\\
5.34999999999999	-0.50449\\
5.39999999999999	-0.504231\\
5.44999999999999	-0.502871\\
5.49999999999999	-0.500484\\
5.54999999999999	-0.497144\\
5.59999999999999	-0.492921\\
5.64999999999999	-0.488449\\
5.69999999999999	-0.484302\\
5.74999999999999	-0.480462\\
5.79999999999999	-0.476909\\
5.84999999999999	-0.474046\\
5.89999999999999	-0.472196\\
5.94999999999999	-0.471225\\
5.99999999999999	-0.471004\\
6.04999999999999	-0.470531\\
6.09999999999999	-0.468938\\
6.14999999999999	-0.466292\\
6.19999999999999	-0.462659\\
6.24999999999999	-0.45813\\
6.29999999999999	-0.452792\\
6.34999999999999	-0.446703\\
6.39999999999999	-0.439919\\
6.44999999999999	-0.432495\\
6.49999999999998	-0.424483\\
6.54999999999998	-0.415935\\
6.59999999999998	-0.406901\\
6.64999999999998	-0.397429\\
6.69999999999998	-0.387565\\
6.74999999999998	-0.377355\\
6.79999999999998	-0.366842\\
6.84999999999998	-0.356122\\
6.89999999999998	-0.34529\\
6.94999999999998	-0.334386\\
6.99999999999998	-0.323447\\
7.04999999999998	-0.31335\\
7.09999999999998	-0.304873\\
7.14999999999998	-0.297909\\
7.19999999999998	-0.292353\\
7.24999999999998	-0.28784\\
7.29999999999998	-0.284033\\
7.34999999999998	-0.280877\\
7.39999999999998	-0.27832\\
7.44999999999998	-0.276308\\
7.49999999999998	-0.274792\\
7.54999999999998	-0.273722\\
7.59999999999998	-0.273053\\
7.64999999999998	-0.272737\\
7.69999999999998	-0.27273\\
7.74999999999998	-0.27299\\
7.79999999999998	-0.273474\\
7.84999999999998	-0.274142\\
7.89999999999998	-0.274957\\
7.94999999999998	-0.275879\\
7.99999999999998	-0.276873\\
8.04999999999998	-0.278273\\
8.09999999999998	-0.280379\\
8.14999999999998	-0.28311\\
8.19999999999998	-0.286388\\
8.24999999999998	-0.289624\\
8.29999999999998	-0.292282\\
8.34999999999998	-0.294364\\
8.39999999999998	-0.29587\\
8.44999999999999	-0.296489\\
8.49999999999999	-0.295952\\
8.54999999999999	-0.294324\\
8.59999999999999	-0.29167\\
8.64999999999999	-0.288929\\
8.69999999999999	-0.286943\\
8.74999999999999	-0.285621\\
8.79999999999999	-0.284878\\
8.84999999999999	-0.2841\\
8.89999999999999	-0.282733\\
8.94999999999999	-0.280778\\
8.99999999999999	-0.278239\\
9.04999999999999	-0.275159\\
9.09999999999999	-0.271581\\
9.15	-0.267506\\
9.2	-0.262934\\
9.25	-0.257864\\
9.3	-0.252298\\
9.35	-0.246237\\
9.4	-0.239684\\
9.45	-0.232643\\
9.5	-0.225118\\
9.55	-0.217112\\
9.6	-0.20863\\
9.65	-0.199675\\
9.7	-0.19025\\
9.75	-0.18036\\
9.8	-0.17001\\
9.85	-0.159211\\
9.90000000000001	-0.147969\\
9.95000000000001	-0.136293\\
10	-0.124188\\
10.05	-0.111659\\
10.1	-0.0987142\\
10.15	-0.0853599\\
10.2	-0.0716046\\
10.25	-0.0571705\\
10.3	-0.0418089\\
10.35	-0.0255713\\
10.4	-0.00850796\\
10.45	0.00924502\\
10.5	0.0275599\\
10.55	0.0464001\\
10.6	0.0657298\\
10.65	0.0855129\\
10.7	0.105714\\
10.75	0.1263\\
10.8	0.147237\\
10.85	0.167672\\
10.9	0.186827\\
10.95	0.20478\\
11	0.22161\\
11.05	0.237698\\
11.1	0.253392\\
11.15	0.268716\\
11.2	0.283695\\
11.25	0.298349\\
11.3	0.3127\\
11.35	0.326769\\
11.4	0.340575\\
11.45	0.353949\\
11.5	0.36674\\
11.55	0.37899\\
11.6	0.390741\\
11.65	0.402362\\
11.7	0.414188\\
11.75	0.426201\\
11.8	0.438387\\
11.85	0.450728\\
11.9	0.463206\\
11.95	0.475806\\
12	0.488515\\
12.05	0.501319\\
12.1	0.514209\\
12.15	0.527168\\
12.2	0.540185\\
12.25	0.553239\\
12.3	0.566315\\
12.35	0.579399\\
12.4	0.592479\\
12.45	0.605537\\
12.5	0.618555\\
12.55	0.631523\\
12.6	0.64443\\
12.65	0.657272\\
12.7	0.670047\\
12.75	0.682743\\
12.8	0.695349\\
12.85	0.707868\\
12.9	0.720298\\
12.95	0.732628\\
13	0.744848\\
13.0500000000001	0.756932\\
13.1000000000001	0.768857\\
13.1500000000001	0.780616\\
13.2000000000001	0.792199\\
13.2500000000001	0.803551\\
13.3000000000001	0.814615\\
13.3500000000001	0.82539\\
13.4000000000001	0.835871\\
13.4500000000001	0.846117\\
13.5000000000001	0.856179\\
13.5500000000001	0.86605\\
13.6000000000001	0.875723\\
13.6500000000001	0.885167\\
13.7000000000001	0.894355\\
13.7500000000001	0.903283\\
13.8000000000001	0.911948\\
13.8500000000001	0.920322\\
13.9000000000001	0.92838\\
13.9500000000001	0.936124\\
14.0000000000001	0.943554\\
14.0500000000001	0.950693\\
14.1000000000001	0.957559\\
14.1500000000001	0.964147\\
14.2000000000001	0.97045\\
14.2500000000001	0.976463\\
14.3000000000001	0.982179\\
14.3500000000001	0.987593\\
14.4000000000001	0.992701\\
14.4500000000001	0.997498\\
14.5000000000001	1.00198\\
14.5500000000001	1.00614\\
14.6000000000001	1.00997\\
14.6500000000001	1.01348\\
14.7000000000001	1.01665\\
14.7500000000001	1.01949\\
14.8000000000001	1.02199\\
14.8500000000001	1.02425\\
14.9000000000001	1.02636\\
14.9500000000001	1.02831\\
15.0000000000001	1.03007\\
15.0500000000001	1.03164\\
15.1000000000001	1.03301\\
15.1500000000001	1.03415\\
15.2000000000001	1.03506\\
15.2500000000001	1.03574\\
15.3000000000001	1.03616\\
15.3500000000001	1.03633\\
15.4000000000001	1.03624\\
15.4500000000001	1.03586\\
15.5000000000001	1.03521\\
15.5500000000001	1.03426\\
15.6000000000001	1.03301\\
15.6500000000001	1.03148\\
15.7000000000001	1.02968\\
15.7500000000001	1.02761\\
15.8000000000001	1.02526\\
15.8500000000001	1.0225\\
15.9000000000001	1.01924\\
15.9500000000001	1.0155\\
16.0000000000001	1.01129\\
16.0500000000001	1.00722\\
16.1000000000001	1.00384\\
16.1500000000001	1.00102\\
16.2000000000001	0.998676\\
16.2500000000001	0.995963\\
16.3000000000001	0.992142\\
16.3500000000001	0.987264\\
16.4000000000001	0.981378\\
16.4500000000001	0.974533\\
16.5000000000001	0.966773\\
16.5500000000001	0.958145\\
16.6000000000001	0.948697\\
16.6500000000001	0.938704\\
16.7000000000001	0.928425\\
16.7500000000001	0.917872\\
16.8000000000001	0.90706\\
16.8500000000001	0.895992\\
16.9000000000001	0.88467\\
16.9500000000001	0.873109\\
17.0000000000001	0.861323\\
17.0500000000001	0.849587\\
17.1000000000001	0.838153\\
17.1500000000001	0.827\\
17.2000000000001	0.816107\\
17.2500000000001	0.805131\\
17.3000000000001	0.793753\\
17.3500000000001	0.781996\\
17.4000000000001	0.76988\\
17.4500000000001	0.757424\\
17.5000000000001	0.744647\\
17.5500000000001	0.731569\\
17.6000000000001	0.71821\\
17.6500000000001	0.704593\\
17.7000000000001	0.690744\\
17.7500000000001	0.67668\\
17.8000000000001	0.662418\\
17.8500000000001	0.647806\\
17.9000000000001	0.632708\\
17.9500000000001	0.617165\\
};
\addplot [color=black, forget plot]
  table[row sep=crcr]{%
6	-1.5\\
6	1.5\\
};
\node[right, align=left, circle,fill=white,draw=black,thick,line width=0.3mm,inner sep=1pt]
at (axis cs:5,0.5) {1};
\addplot [color=black, forget plot]
  table[row sep=crcr]{%
11	-1.5\\
11	1.5\\
};
\node[right, align=left, circle,fill=white,draw=black,thick,line width=0.3mm,inner sep=1pt]
at (axis cs:10,-0.6) {2};
\addplot [color=black, forget plot]
  table[row sep=crcr]{%
16	-1.5\\
16	1.5\\
};
\node[right, align=left, circle,fill=white,draw=black,thick,line width=0.3mm,inner sep=1pt]
at (axis cs:15,-0.6) {3};
\end{axis}
\end{tikzpicture}%
	\end{multicols}
	\vspace*{-0.75cm}
	\caption{
		Evaluation scenario, where the ego vehicle performs a left turn, while maintaining comfortable behavior and courtesy towards other traffic participants. In the first row, a top view of the scene is depicted for three different points in time. The planned trajectory is represented by colored dots where red presents the planned position at $T$. The ego vehicle is red and other vehicles are blue. The center lines of the road lanes are depicted in black. The second row shows the velocity and acceleration of the driven trajectory.
	}
	\label{fig:evalLeftTurn}
\end{figure*}

\section{EVALUATION} 
The evaluation is done using real world map data contained in a high-precision digital map of Ulm (Germany) including lane-changes, intersections, roundabouts and on-ramp scenarios \cite{Kunz2015}. The concept is implemented in C++ using the A*-search algorithm of the DOSL library \cite{Bhattachary2017}. Runtimes are obtained using a Intel XEON E5-1660 v4 CPU with 3.2 GHz utilizing a single thread. 
Other vehicles are simulated with random acceleration uniformly distributed between $[ \SI{-1}{\meter\per\second\squared},\SI{1}{\meter\per\second\squared}]$ in each time step. 
In general, for each of the evaluations about 250 scenarios were analyzed, including lane change, highway on-ramp, roundabout and intersection scenarios.
\subsection{Branching and Heuristic Functions}

At first, the action selection strategy is investigated. 
The corresponding findings are summarized in Table~\ref{tab:MobilEval}. In order to measure the comfort of resulting trajectories, both average squared acceleration $\varnothing a^2$ and jerk $\varnothing \dot{a}^2$ are regarded, as they are also incorporated into the cost function during trajectory planning. The results show that our model-based action selection strategy only has minor influence on the quality of resulting trajectories. This emphasizes that during car-following scenarios exploration of similar states is omitted and that the MOBIL model yields well suited decision making for lane changes when integrated into the graph-based framework.
Thereby, the runtime is reduced by 90\% compared to a more passive action selection strategy similar to the defined preconditions in \cite{Lenz2016}. Consequently, the proposed action selection strategy enables the usage of the extended action set for real-time application, without increased trajectory costs.

{\renewcommand{\arraystretch}{1.0} 
	\begin{table}
		\caption{Evaluation of the action selection strategy, including strategies for car-following and lane changing using the \mbox{MOBIL} model. Thereby, the performance of the proposed action selection strategy is compared with a more passive, i.e. less restrictive, one similar to the preconditions presented in \cite{Lenz2016}. Runtimes for behavior planning as well as average squared jerk and acceleration of driven trajectories are shown. No heuristic functions ($h_0$) are used for the comparison. \vspace{-0.5cm}}
		\label{tab:MobilEval}
		\begin{center}
			\begin{tabular}{c| l l}
				 & \makecell{proposed \\ ($h_0$)}  & passive\\
				\hline
				$\varnothing$ runtime [$\SI{}{\milli\second}$]&\textbf{18.09} & 189.61\\
				max runtime [$\SI{}{\milli\second}$]&\textbf{158.33} & {1588.67}\\
				$\varnothing \dot{a}^2$ $[(\SI{}{\meter}/\SI{}{\second}^3)^2]$&{0.056}& 0.056\\
				$\varnothing a^2$ $[(\SI{}{\meter}/\SI{}{\second}^2)^2]$&{0.39} & {0.39}\\
				\hline
			\end{tabular}
		\end{center}
		\vspace{-0.5cm}
	\end{table}
}

Further, the heuristic functions as well as the overall branching strategy were evaluated, where Table~\ref{tab:IDMEval} shows corresponding results.
As baseline the concepts \cite{Speidel2019,Hubmann2016} are used, which implement a similar branching strategy. It is shown that comfort is much higher for the proposed approach, as $\varnothing a^2$ is slightly reduced and $\varnothing \dot{a}^2$ is nearly 25\% lower. These results emphasize the idea that model knowledge of driver and control models can be effectively used in order to improve comfort, in the trade-off against longer runtimes. However, the proposed heuristic function are able to effectively reduce the runtime. 
It is demonstrated that the average runtime can be reduced by about $17\%$. The maximum runtime is even improved by about $60\%$, from $\SI{158.3}{\milli\second}$ to $\SI{62.67}{\milli\second}$, while the solution of behavior planning remains optimal. Slight divergences of driven trajectories occur due to numerical issues.
As a result, runtimes for behavior planning are significantly reduced even compared to \cite{Speidel2019,Hubmann2016}, while the quality of driven trajectories is improved using the proposed branching strategy.

{\renewcommand{\arraystretch}{1.0} 
	\begin{table}
		\caption{Evaluation of the proposed behavior planning module, where runtimes for behavior planning as well as average squared jerk and acceleration for driven trajectories are shown. The evaluation scenarios were investigated using the proposed branching strategy and the branching strategy implemented in \cite{Hubmann2016, Speidel2019}. Further, the proposed approach is shown with heuristic functions $h_\text{all}$ and without usage of heuristic functions $h_0$. \vspace{-0.5cm}}
		\label{tab:IDMEval}
		\begin{center}
			\begin{tabular}{c| l l l}
				&\makecell{proposed \\ ($h_0$)} & \makecell{proposed \\ ($h_\text{all}$)} & \makecell{\cite{Hubmann2016, Speidel2019} }\\
				\hline
				$\varnothing$ runtime $[\SI{}{\milli\second}]$ &{18.09} &\textbf{14.96} & {14.79}\\
				max runtime $[\SI{}{\milli\second}]$ &158.33 &\textbf{62.67}& 103.00\\
				$\varnothing \dot{a}^2$ $[(\SI{}{\meter}/\SI{}{\second}^3)^2]$&\textbf{0.056} &\textbf{0.055}& {0.073}\\
				$\varnothing a^2$ $[(\SI{}{\meter}/\SI{}{\second}^2)^2]$&\textbf{0.39} &\textbf{0.39} & {0.40}\\
				\hline
			\end{tabular}
		\end{center}
		\vspace{-0.5cm}
	\end{table}
}

\subsection{Motion Planning Framework}
In order to give an insight to the overall performance and the resulting trajectories of the motion planning framework, an exemplary scenario is depicted in Figure~\ref{fig:evalLeftTurn}. In general, an urban left turn scenario is regarded without right-of-way. At time \tikz\node[circle,fill=white,draw=black,thick,inner sep=1pt]{\scriptsize 1};, the ego vehicle \tikz\node[circle,fill=white,draw=red,thick,inner sep=1pt]{\textcolor{red}{\footnotesize e}}; slowly approaches the intersection, while vehicle \tikz\node[circle,fill=white,draw=blue,thick,inner sep=1pt]{\textcolor{blue}{\scriptsize 4}}; crosses it. Afterwards, vehicle \tikz\node[circle,fill=white,draw=blue,thick,inner sep=1pt]{\textcolor{blue}{\scriptsize 2}}; and \tikz\node[circle,fill=white,draw=blue,thick,inner sep=1pt]{\textcolor{blue}{\scriptsize 3}}; approaching from the right have to be considered. Taking the turn in front of vehicle \tikz\node[circle,fill=white,draw=blue,thick,inner sep=1pt]{\textcolor{blue}{\scriptsize 2}}; would cause to much courtesy costs, thus the ego vehicle merges between vehicle \tikz\node[circle,fill=white,draw=blue,thick,inner sep=1pt]{\textcolor{blue}{\scriptsize 2}}; and \tikz\node[circle,fill=white,draw=blue,thick,inner sep=1pt]{\textcolor{blue}{\scriptsize 3}}; at \mbox{time \tikz\node[circle,fill=white,draw=black,thick,inner sep=1pt]{\scriptsize 3};}.

In addition, it is worth noting that during the analysis of all 250 scenarios, the maximum measured overall calculation time for motion planning was $\SI{89.33}{\milli\second}$ using the presented approach. Further, non of the scenarios led to any collisions despite random behavior of other traffic participants. This emphasizes the capability of the framework to handle complex urban scenarios. 

\section{CONCLUSION}
In this work, we presented a motion planning framework for autonomous vehicles in urban environments utilizing graph-search methods. The proposed branching strategy and admissible heuristic functions yield trajectories attributed with lower costs, while the runtime is reduced significantly compared to related work. Therefore, the implementation of the concept on the research vehicle of Ulm University and according validations in real-world public traffic is part of our future work.

\bibliographystyle{IEEEbib}

\end{document}